\documentclass[conference]{IEEEtran}

\usepackage{cite}
\usepackage{url}
\usepackage{graphicx}
\usepackage{amsmath,amssymb,amsfonts}
\usepackage{algorithm}
\usepackage{algpseudocode}
\usepackage{amsmath}
\usepackage{booktabs}
\usepackage{array} 

\algdef{SE}[SUBALG]{Indent}{EndIndent}{}{\algorithmicend\ }%
\algtext*{Indent}
\algtext*{EndIndent}

\graphicspath{ {./images/} }
\usepackage{textcomp}

\begin{document}

\title{Active Rule Mining for Multivariate Anomaly Detection in Radio Access Networks}


 \author{\IEEEauthorblockN{Ebenezer R. H. P. Isaac and Joseph H. R. Isaac}
 \IEEEauthorblockA{
     Global AI Accelerator (GAIA), Ericsson, Chennai, India \\
     \{ebenezer.isaac, joseph.isaac\}@ericsson.com
    }}

\maketitle

\begin{abstract}
Multivariate anomaly detection finds its importance in diverse applications. Despite the existence of many detectors to solve this problem, one cannot simply define why an obtained anomaly inferred by the detector is anomalous. This reasoning is required for network operators to understand the root cause of the anomaly and the remedial action that should be taken to counteract its occurrence. Existing solutions in explainable AI may give cues to features that influence an anomaly, but they do not formulate generalizable rules that can be assessed by a domain expert. Furthermore, not all outliers are anomalous in a business sense. There is an unfulfilled need for a system that can interpret anomalies predicted by a multivariate anomaly detector and map these patterns to actionable rules. This paper aims to fulfill this need by proposing a semi-autonomous anomaly rule miner. The proposed method is applicable to both discrete and time series data and is tailored for radio access network (RAN) anomaly detection use cases. The proposed method is demonstrated in this paper with time series RAN data. 
\end{abstract}


\section{Introduction}
\begin{figure}[b]
	\begin{center}
		\includegraphics[width=\linewidth]{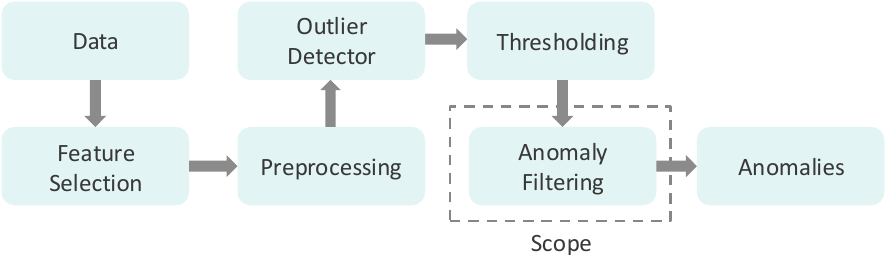}
		\caption{Anomaly Detection Pipeline}\label{fig:ad}
	\end{center}
\end{figure}

Generally, the terms outliers and anomalies are used interchangeably. However, they differ when considered in a business context. An outlier is a rare occurrence that differ significantly from the majority of data under observation. Whereas an anomaly is a significant deviation from the expected occurrence. Not all data points that are outliers conform as anomalies; it depends on the business use case. 
A common anomaly detection (AD) pipeline, as depicted in Fig.~\ref{fig:ad} begins by selecting the features required for the use case. Then the selected feature is preprocessed and passed through an outlier detector which would return the scores. A threshold is set on the scores to flag anomalies. These flags are filtered by some heuristic or rules according to the business case to report the final anomalies.
An AD model would generally imply both the outlier detector and thresholding taken together. In the context of this paper, the output of an AD model refers to the outliers determined after the thresholding step. 
 
The scope of this paper is limited to anomaly filtering step of an AD pipeline. This step involves removal of cases which are not beneficial to the business as false positives and may involve gauging the severity of certain cases for better attention. Anomaly filtering is a trivial task when the input data is univariate. The appropriate action would be to group the occurrences based on statistical similarity and assess them on a case-by-case basis. However, when the data is multivariate, filtering is not straightforward. Multivariate AD algorithms can model complicated anomalies but cannot directly explain why they occur. In a practical scenario, e.g., network KPI (key performance indicator) fault detection, if the reason as to why a certain KPI combination is anomalous is unknown, then the issue causing the anomaly cannot be addressed. Therefore, explainable AD is thus of critical importance. 

Explainers like  SHAP~\cite{lundberg2017unified} and LIME~\cite{ribeiro2016should} depict feature importance for a given prediction. These methods may help understanding which features participate in the occurrence of a given anomaly on a per-case basis. However, they do not formulate generalizable rules that can be assessed by an expert.
For example, in RAN, it is normal for monitoring KPI (such as handover rate) to spike up when load KPI (such as active uplink users) is high. Even if such an occurrence is an outlier in statistical terms, it need not to be considered as a network anomaly when the business objective is to identify faulty behavior of network cells. In a system of multiple load and monitoring KPIs, many of such combinations that does not support the business objective can occur and would be flagged as anomalies by a typical multivariate outlier detection algorithm. An explanator can only show the different levels of importance that each feature has on such occurrences, albeit they are considered as false positives in business context.
There is a knowledge gap between the user of the system and the AD system itself that requires bridging to make its use more fruitful and to enable better automation.

The motivation for the proposed system came from the need for a customer operator to analyse RAN anomalies labelled by our anomaly detection method. However, at that time, there was no efficient way to validate over 2400 anomalies identified from a dataset over 3000 cells considering only 20 performance management (PM) KPIs covering 2 months. We hence invented this method (filed as a US patent by Ericsson~\cite{arm-ad-patent}) to group these anomalies into unique conditions which can then be mapped to actions, i.e., rules. This system helped bring down this number to 126 conditions. The operators found it much easier to assess sample anomalies from each condition and validate them accordingly as opposed to assessing each anomaly one-by-one. This method significantly improved operational efficiency by saving time and reducing human error to validate anomalies. Furthermore, the system paved way to form rules that can be readily implemented in a production system. This paper delineates each stage of this system in detail.

Our proposed anomaly rule miner can be applied to both time series and non-time series datasets, and does not impose any restriction on the underlying outlier detection method or seasonality residual extraction used. The following are the advantages that we claim to be differentiators of our proposed solution:

\begin{enumerate}
	\item The system not only provides a way to explain anomalies but also the means to act upon them.
	\item Rules can be mapped to severity so that anomalies can be prioritized accordingly.
	\item The system also includes the method for handling of unknown rules should they occur.
	\item The system can manage concept drifts as the rules are not based on static thresholds and can evolve based on the domain requirements.
\end{enumerate}$ $

\section{Related Works}
A surrogate model~\cite{molnar2020interpretable} is a decision tree that retrofits features of the dataset with the outputs the AD algorithm as labels. A surrogate model attempts to interpret the AD model based on the combination of the training dataset and its output provided by the AD.\@ Though a standard decision tree can be broken down into a set of rules, the rules obtained from a surrogate decision tree may not depict the complete behavior of the model itself as there may exist a combination of data that is not included as part of the training set. If one were to allow the surrogate decision tree to retrain, the generated rules can possibly vary widely from the previous version. Hence, the ruleset obtained from a surrogate model is not definite.

G. Bruno et al.~\cite{bruno2010mining} consider anomalies as rare association rules that violate dependencies that frequently hold. This work considers the probable occurrence of a concept drift by maintaining an up-to-date set of such functional dependencies. The solution incrementally updates the association rules on append-only databases. Though the system itself is self-learning, it does not give scope to provide feedback to the AD system to solve specific problems that would discount anomalies that are not relevant to the business.

Prathibhamol et al.~\cite{prathibhamol2016anomaly} proposed the use of AD and association-rule mining as a means for multilabel classification. The method involves clustering the data first using k-means clustering, followed by oversampling PCA (Principal Component Analysis) for anomaly detection, then frequent itemset matching to associate class labels. Though the solution involves the use of unsupervised learning techniques, the scope is limited to supervised learning.



Sivapalan et al.~\cite{sivapalan2023interpretable} discusses a novel,
explainable rule-mining approach for ECG anomaly detection. This work combines
artificial neural networks (ANN) with a rule-based system to classify normal and
abnormal heartbeats. The solution is highly accurate, and
suitable for real-time IoT-enabled wearable sensors. It achieves over 90\%
sensitivity and accuracy, minimizing power consumption by only transmitting
only the abnormal heartbeats and hence saving energy.

Heinrich et al.~\cite{heinrich2023rule} made a system to detect cyberattacks on
railway signaling. It focuses on defending against a Dolev-Yao attacker who can
perform semantic attacks. This work utilizes a distributed, rule-based anomaly
detection system, and ensures commands are validated against the real-time state
of neighboring railway elements, detecting all attacks without false positives.
The system introduces minimal latency and is scalable for real-world railway
systems. However, a small overhead is introduced by enabling Field Elements
(FEs) to communicate with their neighboring FEs within the railway topology.

The literature, mentioned so far, do not appreciate the importance of business-specific feedback that can be provided to the system. The rules incorporated in those use cases are not extensible. These solutions not designed for a system than can evolve as per the requirements of the network and the business case. For a given dataset, it is possible to identify multiple anomalies that would be of importance to one problem scope but not another. As our understerstanding of the network improve, so should these rules. Nevertheless, there are also studies in literature that considers active anomaly detection that considers human-in-the-loop feedback to improve prediction accuracy. Some of such studies are summarized in Jari’s thesis~\cite{jaaskela2020anomaly}.

Kun Liu et al.~\cite{liu2021interactive} adequately considers the expert feedback to tune AD within an information system. In here, the rules are mined for the behavior of the system based on historical data and those rules are evaluated by the expert. The method considers both frequency of occurrence of these rules through frequent itemset matching and their associated behavior. However, their rules are analogous to simple first-order logic and the solution does not discuss the handling of real values such as that of a network KPI.\@

Steenwinckel et al.~\cite{steenwinckel2021flags} have designed a generic framework to combine AD, fault recognition and root cause analysis. The design is also considered to be context-aware, adaptive, and interpretable. It claims to combine unsupervised AD with supervised labels provided by an expert. It does not mandate the algorithms used for each component, only that the interpretations are stored in a knowledge graph. It did not discuss how to handle multiple contexts or concept drifts.

A detailed survey of over 150 articles on explainable anomaly detection (XAD) methods is provided by Li et al.~\cite{li2023survey}. They postulate that the oracle definition of an anomaly is one that is provided by the end-users of the system based on application-specific domain knowledge. In this definition, they also mention that \textit{true anomalies} strongly depends on real-world context and are often hard to formally or precisely define. Hence, no universal definition of an anomaly exists. An XAD method aims to explain why a given anomaly detector finds a certain instance to be anomalous. In their study, the state that there is no standard method to evaluate XAD techniques. They further mention that most existing XAD methods, such as SHAP, have a high computational cost that limits their scalability for large data applications. 

\begin{figure}
	\begin{center}
		\includegraphics[width=\linewidth]{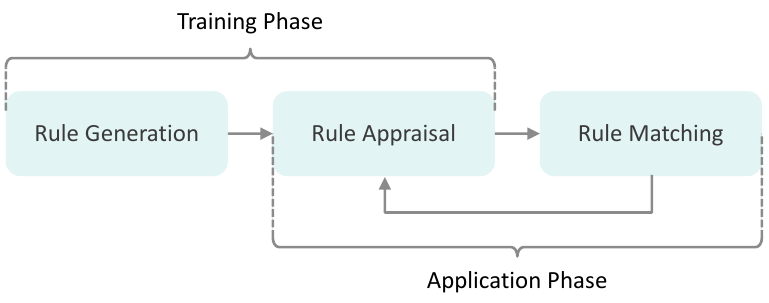}
		\caption{Phases and Stages of the Anomaly Rule Miner}\label{fig:activerulemining}
	\end{center}
\end{figure}

\section{Active Rule Mining – An Overview}
The proposed method defines active rule mining as the process of identifying unique conditions that can be associated with actions to form prosective rules to define anomalies. The first step is to group instances, labelled as anomalies by the anomaly detector, according to their statistical characteristics with reference to the ones that are not anomalous. Then, map these groups to  condition vectors that could be readily understood by a human expert. The system also mentions the number of anomalous occurrences of each group within the training set.

Consider this example: given an anomalous record, let $x_1$, $x_2$, and $x_3$ be the observed values of fields $X_1$, $X_2$, and $X_3$ respectively. If the statistical measure for comparison is the median, then a sample condition vector may be 
\[[x_1  < \textnormal{median}(X_1), ~x_2  > \textnormal{median}(X_2),~ x_3 \approx \textnormal{median}(X_3)]\]
where the median is calculated from the subset of data that is not labelled anomalous by the outlier detector. The statistical measures are periodically updated to counteract possible concept drift. Note that the system also expects the outlier detector to be equipped to handle concept drifts, e.g., by incorporating the method prescribed by Isaac and Sharma~\cite{ath-paper}.
The expert can then provide feedback to the system for each vector to form a rule. The feedback can include accepting the condition to be anomalous and assign an appropriate response or whitelisting a condition as not anomalous.

The solution also allows the rules to be combined or split based on business need. An example of a combined rule in the above case can be to consider all occurrences where the first two conditions of $x_1$ and $x_2$ apply regardless of the value of $x_3$. Splitting a rule would mean to consider a single generated rule as two or more separate occurrences pertaining to a set of conditions. For instance, in the above case, conditions 1 and 3 can be part of one rule and conditions 2 and 3 may form another rule.

Mapping anomalies to rules associates anomalies with necessary response actions. Each rule corresponds to actions an operator should take if that anomaly occurs. The outlier detector is still used to discover new anomalies not present in training data. When an anomaly that does not correspond to an existing rule set have been discovered by the rule miner system, it creates the rule and adds to the list of rules that have not yet been appraised by an expert. Unappraised anomalies raise alarms by default due to their unknown nature, while appraised anomalies have defined rule-based actions.

\section{Method}
The system is composed of three stages: (1) Rule Generation, (2) Rule Appraisal, and (3) Rule Matching.
The solution connects the above stages into the training and application phases, as shown in Fig.~\ref{fig:activerulemining}. The appraisal is a common component for both phases as during rule matching, it is possible to discover new rules, which would also require an appraisal from an expert. 

\subsubsection{Training Phase}
The training phase starts with the rule generation stage, which takes in the output of the outlier detection algorithm (after thresholding, if any) to formulate rules, i.e., condition vectors with an action (initially an empty action). These rules are passed on to a rule appraisal stage, where they are assessed by a subject matter expert (SME) to confirm which of these rules would provide value to the business and possibly include corresponding responses to their occurrence.

\subsubsection{Application Phase}
Once an outlier is detected in production, in the rule-matching stage, the system searches for the associated rule for the occurrence. Once a match is found, a corresponding action can be taken based on the response mapping in the rule. If there is an anomalous occurrence that does not match the appraised rule set, then a corresponding rule is created with a default action to be appraised by the SME again in the rule appraisal stage.

\begin{figure*}
	\begin{center}
		\includegraphics[width=0.9\linewidth]{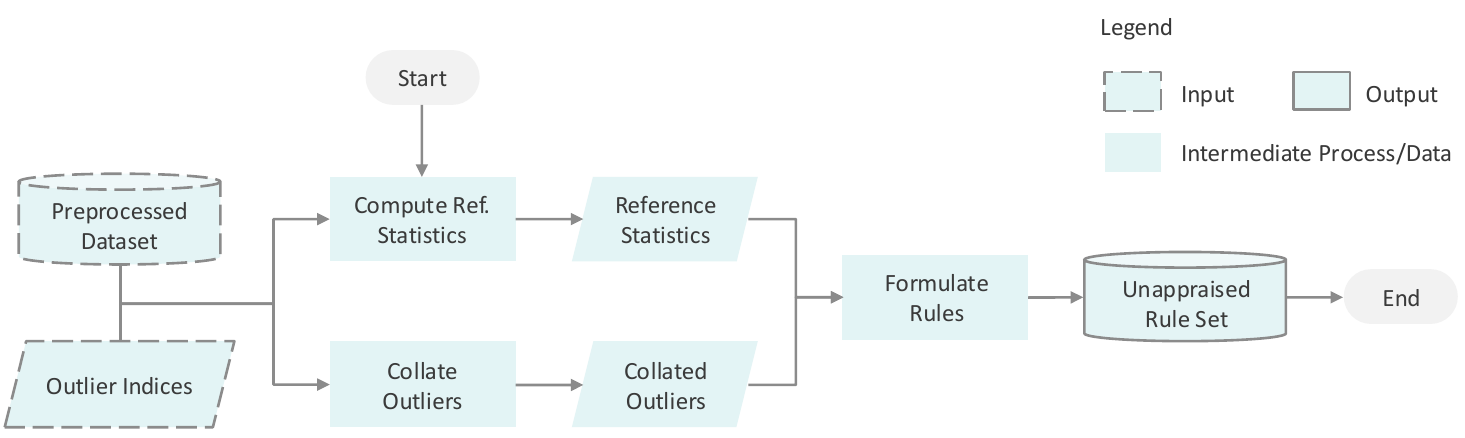}
		\caption{Rule Generation Flow 
		}\label{fig:rulegen}
	\end{center}
\end{figure*}

\subsection{Rule Generation Stage}
The training dataset is preprocessed and has its outliers labeled according to the steps prior to anomaly filtering in the AD pipeline (Fig.~\ref{fig:ad}). The rule generation stage, as depicted in Fig.~\ref{fig:rulegen}, requires the preprocessed training dataset and the indices of the outliers as input. These indices are timestamps in the case of time series AD. In this stage, the system compares each anomalous occurrence with a reference statistic of its context to formulate rules that can be appraised by an expert.

\subsubsection{Compute Reference Statistics}\label{sec:computereference}
A reference statistic is based on two components: context and statistical measure.

    \paragraph{\textbf{Context}} a set of instances, $X$, that have similar circumstances. An outlier is determined in terms of its context. In RAN cell KPIs, the context may be divided according to three levels: region level, cell level, and KPI level. Iterating through contexts would mean to go over all combinations of desired level of context. E.g.,
          \begin{itemize}
              \item KPI-level context: set of all values of a given KPI over a specific period
              \item Cell-KPI-level context: set of all values of a given KPI over a specific cell and period
              \item Region-KPI-level context: set of all values of a given KPI over all cells of a specific region and period
          \end{itemize}
          Note that the time window or period is always part of a context and is configurable.
          
    \paragraph{\textbf{Statistical measure}} the calculation that is applied on the context to define a norm. An outlier is a deviation from this norm. Let $X$ be a random variable that denotes the values within a context. Statistical measures can be simple like $mean(X)$, $median(X)$, or $mode(X)$. The measure can also be a combination of any aggregate such as
          \begin{itemize}
              \item $mean({x | x \in X, x > median(X)})$, the mean of all values that are greater than the median
              \item $mean(X) + std(X)$, the sum of the mean and standard deviation of all values
          \end{itemize}
          A reference statistic is computed for each context using the statistical measure. E.g., for a cell-KPI level context with $n$ KPIs and $m$ cells, the following would be the set of reference statistics computed:
			
		  \vspace{0.5em}
          \begin{tabular}{llll}
              $\text{cell}_1\text{kpi}_1\text{ref}$, & $\text{cell}_1\text{kpi}_2\text{ref}$, & \ldots & $\text{cell}_1\text{kpi}_n\text{ref}$ \\
              $\text{cell}_2\text{kpi}_1\text{ref}$, & $\text{cell}_2\text{kpi}_2\text{ref}$, & \ldots & $\text{cell}_2\text{kpi}_n\text{ref}$ \\
              $\text{cell}_3\text{kpi}_1\text{ref}$, & $\text{cell}_3\text{kpi}_2\text{ref}$, & \ldots & $\text{cell}_3\text{kpi}_n\text{ref}$ \\
              \ldots                                 & \ldots                                 & \ldots & \ldots                                \\
              $\text{cell}_m\text{kpi}_1\text{ref}$, & $\text{cell}_m\text{kpi}_2\text{ref}$, & \ldots & $\text{cell}_m\text{kpi}_n\text{ref}$ \\
          \end{tabular}
          \vspace{0.5em}

\begin{figure*}
    \begin{center}
        \includegraphics[width=0.9\linewidth]{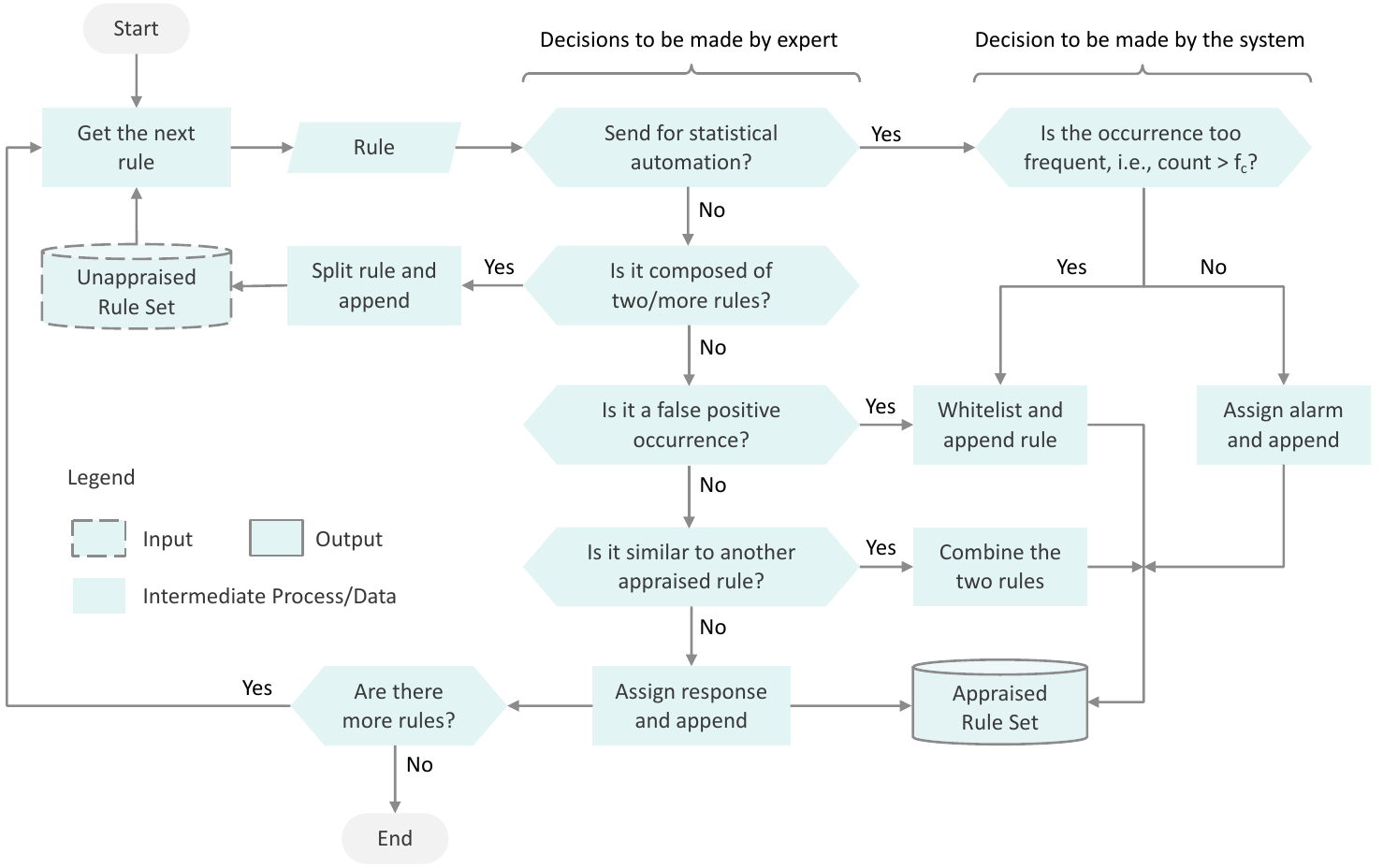}
        \caption{Rule Appraisal Flow. Once the rules are generated, they are appraised by a domain expert. A rule can be split to multiple rules, combined with an existing appraised rule, whitelisted as a false positive, or assigned an appropriate response to be taken on its occurrence. The rule may also be whitelisted if its count in the dataset is beyond the critical frequency, $f_c$, for statistical automation.
        }\label{fig:ruleappraisal}
    \end{center}
\end{figure*}

\subsubsection{Collate Outliers}\label{sec:collateoutliers}
This is a process that only applies to time series data wherein a set of consecutive outlier instances can be grouped together and be considered a single anomalous occurrence with the start and end times specified for reference. A sequence of outliers can be considered consecutive if they have a configurable minimal interval between them. That is, the interval between each succeeding outlier does not exceed a specified period, e.g., 15 minutes. Once consecutive outliers are identified, the fields should be aggregated to condense the records of each consecutive outlier sequence to a single record.

Let each outlier record be represented as $[t,x_1,x_2,x_3,..x_n]$ where $n$ is the number of fields in the preprocessed data and $t$ is the timestamp. A consecutive set of outliers, $C$, can be collated as a single record as
\begin{multline*}
    b = [t_s,t_e,\text{agg}_1 (\text{all}~x_1~in~C),\text{agg}_2 (\text{all}~x_2~in~C), \\
    \ldots,\text{agg}_n (\text{all}~x_n~in~C),\text{duration}]
\end{multline*}

where $t_s$ is the start time of the sequence, $t_e$ is the end time of the sequence, and $agg_i ()$ can be any aggregation function (just like the statistical measure in the previous section) for the $i^{th}$ field. Typical choices may be mean, maximum, and minimum. The designer may keep all aggregates to be the same or have different for each field. E.g., maximum aggregation for monitoring KPI and minimum aggregation for load KPI.\@ Duration is a derived attribute that may either indicate the difference between $t_e$ and $t_s$, or the number of records in $C$, i.e., the number of outliers in the sequence. The duration field can be a useful component to understand the severity of the anomaly.

\subsubsection{Formulate Rules}\label{sec:formulate rules}
A rule is comprised of at least 3 parts: a condition vector, the count of occurrence and a response. The condition vector is a set of conditions that compare the observed value of each field of the collated outlier record to the corresponding reference statistic in its context. An observed value can only be considered comparatively greater or lesser than the reference value if the difference is significant.

A rule can hence be represented as follows.
    \[[\text{condition}_1,\text{condition}_2,\ldots \text{condition}_n,\text{count},\text{response}]\]
where the $\text{condition}_i$ indicates whether observed value of the $i^{th}$ field is either greater, lesser, or approximately equal to its reference statistic. The count field indicates how many times the outliers that pertains to all these conditions have occurred in the training data. The response field will be populated in the rule appraisal stage. It indicates what should be done as a response to the occurrence of an anomaly that satisfies these conditions. The response field may include multiple nested fields as required to store relevant information such as severity, priority, type, etc.\@ in addition to the sequence of actions to be performed.

A condition can also be applied on the duration field for time series data as well with a user-defined constant as a reference statistic. E.g., duration $>$ 4, would mean if the anomaly has occurred consecutively greater than 4 times.

The Algorithm~\ref{alg} below shows a possible way of how the rule formulation can be achieved. The checks in step 2.2.2 and 2.2.3 (whether  $x_i \gg ref$, or $x_i \ll ref$) can be done based on a heuristic that is suitable for the data. E.g., if the concerned field varies on a linear scale, then a simple difference threshold can be used.
\begin{equation}
    x_i - \text{ref} > \theta \Rightarrow  x_i \gg ref
\end{equation}
\begin{equation}
    x_i - \text{ref} < -\theta  \Rightarrow  x_i \ll ref
\end{equation}

If the field varies at an exponential scale, then a ratio-based threshold may be used
\begin{equation}
    \frac{x_i}{\text{ref}} > \theta \Rightarrow x_i \gg ref
\end{equation}
\begin{equation}
    \frac{\text{ref}}{x_i} > \theta \Rightarrow  x_i \ll ref
\end{equation}

The condition check may be computed as a combination of difference and ratio-based methods. The rule miner mandates the existence of the check to show how the observed anomalous record compares to the reference statistic, but not the way it is implemented.

 \begin{algorithm}[t]
	\caption{Formulate Rules} 	\label{alg}
	
	\small\vspace{0.2em}
	\begin{enumerate}
		\item[1.] Start by initiating an empty unappraised rule set
		\item[2.] For each collated outlier record
		\begin{enumerate}
			\item[2.1] Initiate empty rule vector
			\item[2.2] For each field $x_i$ in the record
			\begin{enumerate}
				\item[2.2.1] Let ref be the corresponding reference statistic in the outlier’s context
				\item[2.2.2] If $x_i \gg ref$, then, append to rule vector the value 1 to indicate that $x_i > ref$
				\item[2.2.3] If $x_i \ll ref$, then, append to rule vector the value -1 to indicate that $x_i < ref$
				\item[2.2.4] Otherwise, append to rule vector the value 0 to indicate that $x_i \approx ref$
			\end{enumerate}
			\item[2.3] If the rule vector already exists in the unappraised rule set, then increment the count by one
			\item[2.4] Otherwise, add the rule vector to the unappraised rule set
		\end{enumerate}
		\item[3.] End by returning the unappraised rule set
	\end{enumerate}
\end{algorithm}

Step 2.3 shows how similar outliers are in the same rule. This step would identify that the current outlier occurrence matches an existing rule and increments its count. Therefore, all rules in the resulting unappraised rule set are unique, with a count for each rule.

\begin{figure*}
	\begin{center}
		\includegraphics[width=0.9\linewidth]{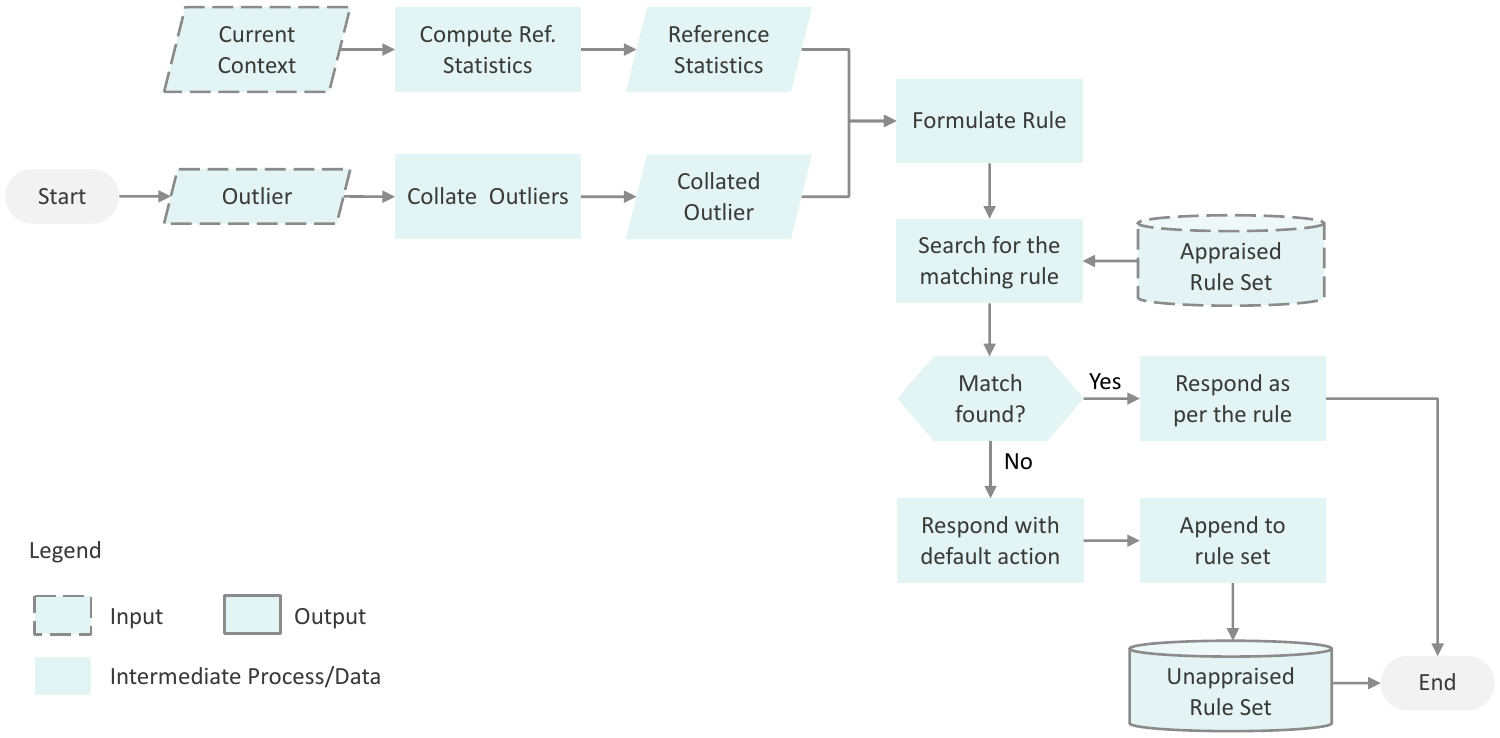}
		\caption{Rule Matching and Discovery Flow. In production, each anomaly is tested against the rule set.  Once a matching rule is found, appropriate action is taken based on the response set during appraisal. However, if a matching rule is not found, a default action is taken and then a new rule is created corresponding to the anomalous occurrence.}\label{fig:rulematch}
	\end{center}
\end{figure*}

\subsection{Rule Appraisal Stage}\label{sec:ruleappraisal}

The unappraised ruleset is assessed one-by-one by SME who then defines appropriate responses for the anomalous occurrence. The overall flow of this stage is depicted in Fig.~\ref{fig:ruleappraisal}.
For each rule, an SME can take one of four actions: assign response, split, combine, or whitelist. Note that all rules that are appraised are then appended to the appraised rule set.

\subsubsection{Assign Response}
The response can be an action or a sequence of actions to be followed should an anomaly matching this rule occurs. It can be simple as alerting with an alarm, creating an automated ticket, or even taking remedial actions as part of close loop automations to counteract the anomaly. E.g., in a thermal power plant, an anomalous increase in pressure and temperature can be acted upon by reducing the fuel feed and initiating the cooling process.

The count variable serves to support the SME to assign an appropriate response. The SME may also add custom conditions to extend the rule, e.g., $x_1 > 2x_3$, duration $>$ 5 intervals, etc.

\subsubsection{Split rule}
If the rule can be considered a composition of two or more rules to which individual actions can be assigned, it can be split accordingly. The rules that are split are added back to the unappraised rule set to be assessed separately.
E.g., If a rule is represented as a numerical vector as shown in Section~\ref{sec:formulate rules},  some of the possible ways a condition sequence of a rule [a, b, c, d, e, f, g] may be split can be

\begin{itemize}
    \item~[a, b, c, d, x, x, x] and [x, x, x, x, e, f, g]
    \item~[a, b, x, x, e, f, g] and [a, b, c, d, x, x, x]
    \item~[a, x, c, x, e, x, g] and [x, b, x, d, x, f, x]
\end{itemize}
assuming letters `a' through `g' refers to conditions and `x' stands for a ``don’t-care'' condition (similar in definition to a ``don't-care'' condition in digital systems). To simply put, `x' can assume any condition. E.g., the rule [a, b, c, d, x, x, x] can be considered satisfied if just conditions a, b, c, and d, are true regardless of the other values specified by `x'.

Note that once the rules are split with ``don’t care'' conditions, their occurrence count may increase due to other matching rules and so their count variable will be recomputed accordingly.

\subsubsection{Combine rule}
If the rule is similar to another previously observed and appraised rule and can be associated with the same response, then this rule can be combined with the corresponding appraised rule.
E.g., the rules with conditions [a, b, c, d$_1$, \ldots] and [a, b, c, d$_2$, \ldots] can be combined as [a, b, c, x, \ldots] with a common response. Where d$_1$ and d$_2$ may indicate two different conditions but do not affect the overall definition of the rule according to the SME.\@

\subsubsection{Whitelist rule}
There are rules that may look like significant outliers to the AD algorithm, but their occurrence may not impact the business. Hence, such rules that are not of business importance can be whitelisted as false positives and no action needs to be taken for their occurrence. The simplest way to whitelist a rule would be to associate its condition sequence with a null response.

By definition, a condition that occurs quite often, i.e., with high count, may need not be considered an anomaly at all. Such rules may optionally be whitelisted as not anomalous. Thus, the user can set a critical frequency, $f_c$, and choose to whitelist all rules with a count above this frequency. The remaining rules can be associated with a default alarm. Although this may not be the most effective way to handle anomalies, this process can increase efficiency through automation if SME availability is limited.

\begin{figure*}
	\begin{center}
		\includegraphics[width=0.85\linewidth]{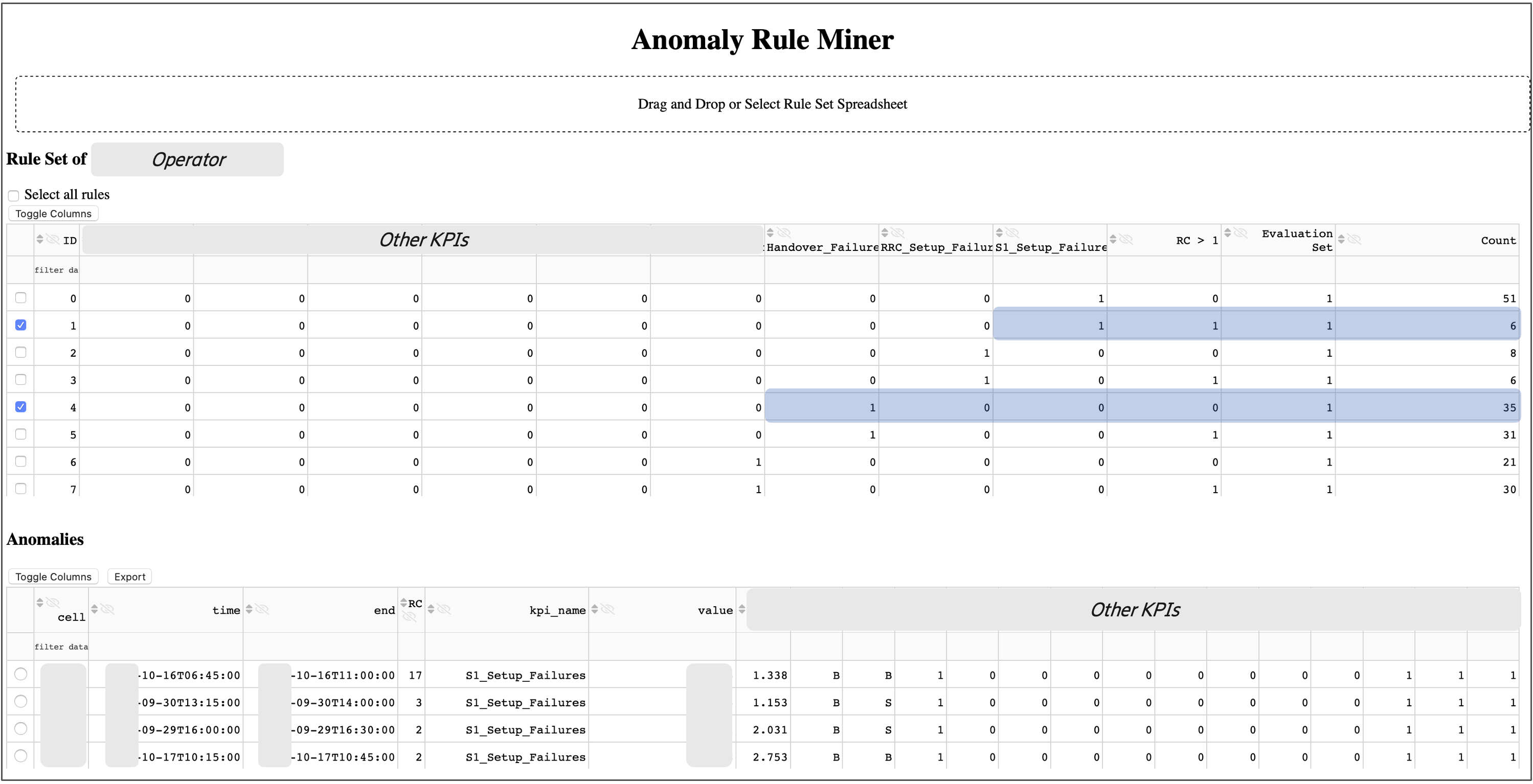}
		\caption{Rule Appraisal User Interface Prototype}\label{fig:rule-appraisal-ui}
	\end{center}
\end{figure*}

\subsection{Rule Matching Stage}
Once the system is deployed, it will actively check the condition of every outlier detected to see if it matches a rule in the set of appraised rules as shown in Fig.~\ref{fig:rulematch}. If a matching rule is found, the system follows the response associated with the rule. If not, the system creates a new rule corresponding to the unseen occurrence and appends it to the unappraised rule set to be appraised by an SME.\@

\subsubsection{Compute Rule}\label{sec:computerule}
This step reuses many of the components in the rule generation stage. The difference being the rule generation stage works on the training dataset to produce a rule set whereas the rule matching stage works on the current context to evaluate live anomalies in the production environment.

The system generates a set of conditions that reflect the observed outlier. For this step, the system requires to persist the current context and compare it to the configured statistical measure described in Section~\ref{sec:computereference}. Note that the current context will be recomputed periodically to deal with possible concept drifts. In the context of this paper, a concept drift is a significant change in the behavior of the system’s environment which can cause an increase/decrease in typical values of the variables monitored by the system. E.g., when a new office building is opened, the load KPIs of the cell towers in the vicinity would all exhibit an overall peak in their normal working conditions. Such behavior will be accounted by the system by keeping the statistics of the context up to date. The periodicity of the training and the time window of the context depends on the problem. In general practice, the system can keep the context of a 30-day time window and update with a daily periodicity.

The outlier here can also be a collated outlier in the case of time series data. Each time an outlier occurs within the minimum interval, it is collated with the most recent outlier by increasing the duration by one and updating its fields according to the steps mentioned in the Collate Outliers Section~\ref{sec:collateoutliers}. The system can be configured to do one of two options.


    \paragraph{Delayed notification option} Repeat the collate outliers process until no more outliers can be collated, i.e., the minimum interval has elapsed. Then, send a single collated outlier to the formulate rule process. This option will ensure that there would be no multiple anomaly flags for a single consecutive set of outliers. However, the anomaly will only be notified by the system once the entire duration of the anomalous occurrence has elapsed.
    
    \paragraph{Eager notification option} For every outlier, collate with the most recent outlier (if within minimum interval), then send it to the formulate rule process. This may create multiple anomaly flags for each consecutive outlier. However, this option is more useful to be notified of the anomaly before it becomes too severe or when the anomaly is prolonged.


\subsubsection{Search for Matching Rule and Respond}
The formulated rule pertaining to the observed outlier is compared to the rules of the appraised rule set. Once a matching rule is found, the steps as mentioned in the response field of the matching rule is executed by the system.

Since every outlier ought to be mapped to a response. The system is configured with a default action that it must take if it experiences an outlier that does not match the rules of the appraised rule set. After the action, it uses the conditions created by the previous step (Section~\ref{sec:computerule}) to formulate a rule following the steps in Section~\ref{sec:formulate rules} and adds it to the unappraised rule set. If there is already a matching rule in the unappraised rule set, then the system, instead of appending the rule, would increase the count of the existing rule by one.

In the application phase, an SME is expected to periodically follow the rule appraisal stage (Section~\ref{sec:ruleappraisal}) should there be unappraised rules observed by the system in rule matching stage.

\begin{figure}
	\begin{center}
		\includegraphics[width=\linewidth]{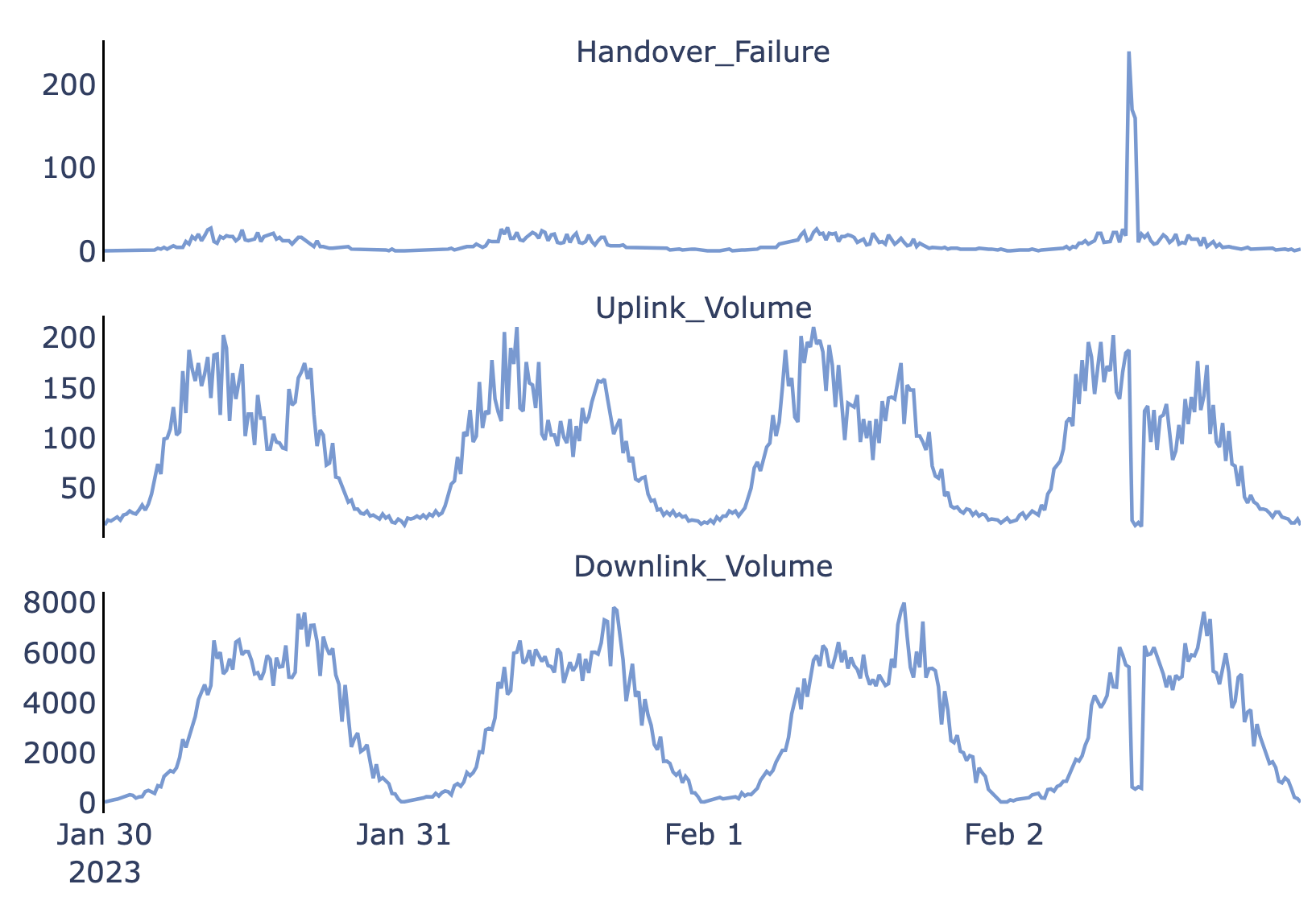}
		\caption{Example of a significant anomalous condition}\label{fig:anomaly-example}
	\end{center}
\end{figure}

\section{Examples and Discussion}
Initially, the multivariate AD system, as depicted in  Fig.~\ref{fig:ad}, is to be executed without filtering to gather indices that the outlier detector with thresholding predicts as outliers. This is the setup process to gather the inputs necessary for the rule generation stage. The preprocessed dataset produced by the preprocessing step of the AD pipeline and the outlier indices are fed to the rule generation stage to produce the unappraised rule set. In our case, it was a spreadsheet for an easier reference for both computers and humans.

Fig.~\ref{fig:rule-appraisal-ui} shows a simple prototype of the rule appraisal interface. The user can check the rules of interest in the Rule Set section (two rules are selected in the example) and the outliers corresponding to the selected rules will be listed in the Anomalies section. The user can view each anomalous instance by clicking on the radio button to observe the graph for insights and then proceed by assigning the action for the rule such that it will apply to all occurrences that correspond to the same pattern. 

\begin{figure}
	\begin{center}
		\includegraphics[width=\linewidth]{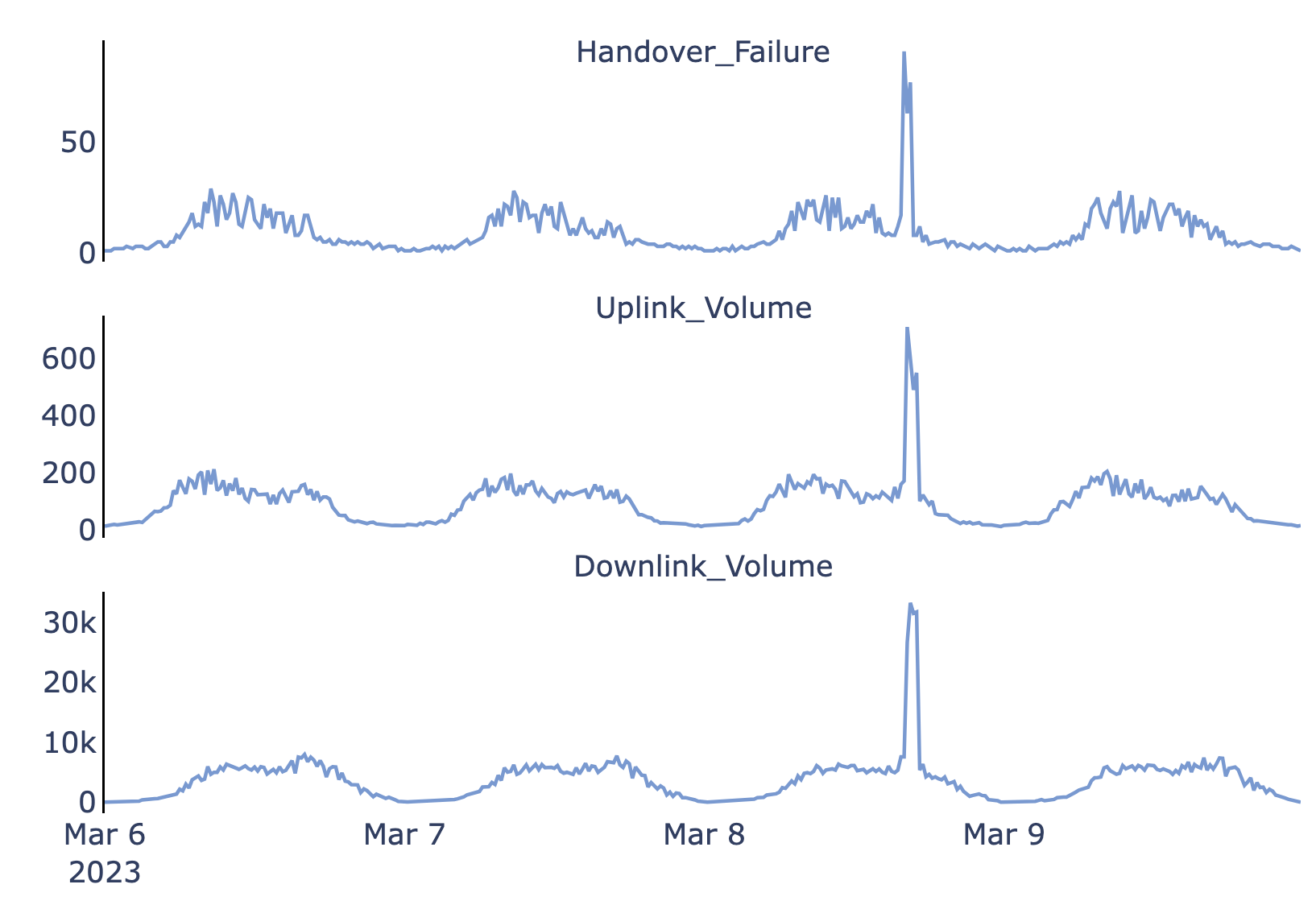}
		\caption{Example of a whitelist condition}\label{fig:whitelist-example}
	\end{center}
\end{figure}

 Fig.~\ref{fig:anomaly-example} depicts an example of a significant anomalous combination of business interest. The peak in handover failure has caused a dip in uplink and downlink volume. A peak in handover failures suggests that the cell is experiencing difficulties in successfully handing over connections to neighboring cells. The dip in both downlink and uplink volume indicates that the overall data traffic passing through the cell has decreased as a result. This event could mean that there is a an issue worth addressing from the operator's perspective since it directly related to customer experience. This may be due to  problems with the cell's coverage, such as interference, weak signal strength, or cell overlap issues that can be caused by a change in the cell configuration parameters. In Fig.~\ref{fig:whitelist-example}, there is a peak in all three: handover failure, uplink volume and downlink volume. It is natural for handover failures to occur when the load is unusually high. It can happen when there is an event that caused a spike in traffic beyond usual levels, e.g., a football match or a riot. Such events are outside the control of the network operator. Statistically, this is a signficant outlier combination, but from a business perspective, it is an expected occurrence where the cell is behaving as it is supposed to. Hence, this condition is whitelisted in the appraised rule set.

Once all the rules are appraised, the system is put through production where each anomalous occurrence is pitted against these rules. If any new combination occurs, it is added to the unappraised rule set with a notification to the network operator to ascertain the corresponding action beyond a generic notification. 

Network operators troubleshoot alarms from a network of tens of thousands to hundreds of thousands of cells. Each cells have hundreds of PM KPIs to assess. Consider for a small case of 3000 cells and an assessment of just 20 KPIs. Assume a typical threshold-based anomaly detector that raises at least 10 anomalies per cell every month. That would mean 30,000 anomalies a month for the operator to analyze and decide to take corresponding action. Since AI-based anomaly detectors are more sensitive to capturing rare patterns in data, operators have shared that anomalies detected by such systems increase this count by 30\%. It is imperative for operators to prioritize the most impactful anomalies and handle only what is under their control. 

\section{Conclusion}
This paper proposes an active rule mining approach that bridges the gap between multivariate anomaly detection systems and domain experts' knowledge. The method generates interpretable anomalous conditions from detected anomalies, allowing experts to assess and assign appropriate actionable responses to form rules. These rules enables the system to adapt to business contexts. The solution supports both time series and non-time-series data, and handles concept drifts by periodically updating reference statistics. Future prospects may include extending this system beyond RAN to vet its applicability to other parts of the telco stack such as core networks.

\bibliographystyle{IEEEtran}
\bibliography{ARM-AD}
\end{document}